# A Bootstrap Machine Learning Approach to Identify Rare Disease Patients from Electronic Health Records

Ravi Garg, MSc[1], Shu Dong, MS[1], Sanjiv Shah, MD[2], Siddhartha R Jonnalagadda, PhD[1]
[1]Division of Health and Biomedical Informatics, Department of Preventive Medicine, Northwestern University Feinberg School of Medicine, Chicago, IL
[2]Division of Cardiology, Department of Medicine, Northwestern University Feinberg School of Medicine, Chicago, IL

Abstract

*Rare diseases are very difficult to identify among large number of other possible diagnoses. Better availability of patient data and improvement in machine learning algorithms empower us to tackle this problem computationally. In this paper, we target one such rare disease – cardiac amyloidosis. We aim to automate the process of identifying potential cardiac amyloidosis patients with the help of machine learning algorithms and also learn most predictive factors. With the help of experienced cardiologists, we prepared a gold standard with 73 positive (cardiac amyloidosis) and 197 negative instances. We achieved high average cross-validation F1 score of 0.98 using an ensemble machine learning classifier. Some of the predictive variables were: Age and Diagnosis of cardiac arrest, chest pain, congestive heart failure, hypertension, prim open angle glaucoma, and shoulder arthritis. Further studies are needed to validate the accuracy of the system across an entire health system and its generalizability for other diseases.*

## Introduction

Because of several well-established factors such as a) subjectivity involved in human decision-making, b) domain knowledge, and c) prior experience,[1,2] there is a possibility of type-1 and type-2 errors in identifying whether patients satisfy a known set of inclusion and exclusion criteria. Furthermore, clinicians typically rely on patterns learnt from patients encountered in their own settings, thereby missing out on the advantage of screening patients or diagnosing them based on latent knowledge and patterns present across an entire health system. An automated process for prescreening and early diagnosis would be quicker and serve as an independent judge. Rare or orphan diseases defined by the Rare Diseases Act of 2002 as diseases that affect fewer than 200,000 people in the United States (equivalent to approximately 6.5 patients per 10,000 inhabitants[3]) present a special case of this. There are about 8,000 rare diseases such as cardiac amyloidosis today and they appear early in life often proving to be fatal or chronically debilitating.[4,5] Cardiac amyloidosis is caused by abnormal deposits of protein called amyloid in the heart tissue.[6] These tissues make it hard for the heart to work properly. It is very difficult to successfully recognize a potential cardiac amyloidosis patient manually among large number of other patients. For enrolling rare disease patients for clinical trials, clinicians usually assign pre-defined constraints to filter out most negative patients. However, they still have to manually go through the remaining large part of each patient's medical record to precisely arrive at a correct decision. For decision making, clinicians have a higher chance of missing the diagnosis of a rare disease since the symptoms are largely unknown for many rare diseases and such patients are encountered only a few times during normal care.

The creation and adoption of electronic health records (EHRs) ignited widespread interest and created abundant opportunities for clinical and translational research.[7] As Friedman et al noted, the extensive use of clinical data provides great potential to transform our healthcare system into a "Self-learning Health System."[8,9] In addition to its primary purpose of providing improved clinical practice, the use of EHRs offers means for the identification of participants who satisfy predefined criteria. This can be used for a variety of applications, including clinical trial recruitment, outcome prediction, survival analysis, and other retrospective studies.[10-13] The overall goal of this research is to develop a platform that identifies patients potentially with any rare disease or has the risk of any risk disease. We are testing this approach initially on cardiac amyloidosis; however, our approach is scalable to all rare diseases. The main components of our platform are to represent each patient as a multidimensional vector and then to use machine-learning algorithms based on available data on rare diseases to automatically identify them.

**Methods**

Figure 1 describes the architecture of our approach.

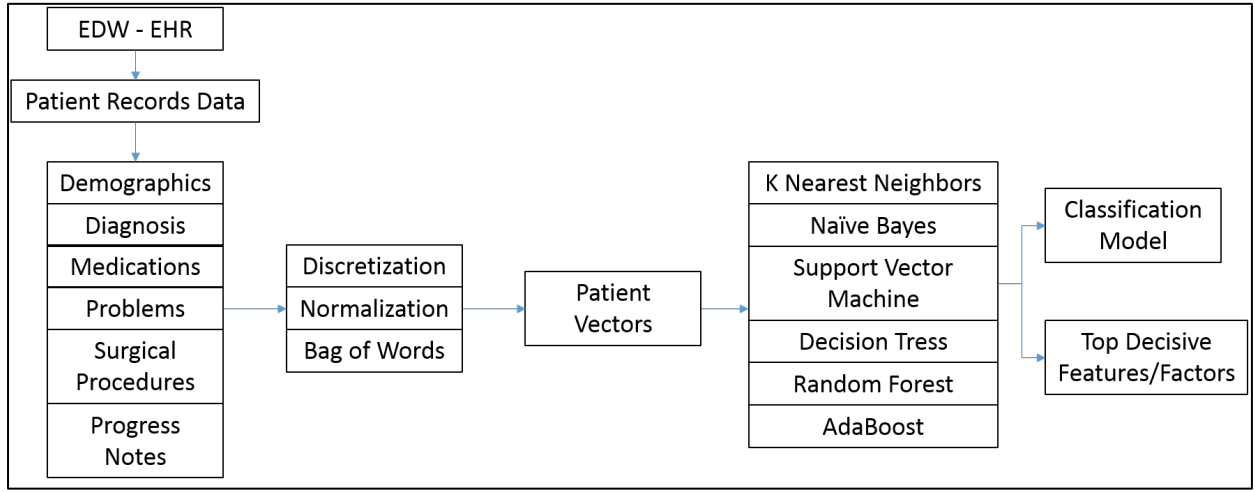

**Figure 1: Architecture of our system.** We first collect patient records from the Northwestern Medicine Enterprise Data Warehouse (NMEDW). Next, the important sections of patient records are extracted and transformed to obtain patient vectors. These vectors are then used to train machine learning algorithms which gives the final classification model and a list of top decisive features.

First, we created a reasonable sized dataset of positive and negative cardiac amyloidosis patients using information from past and ongoing clinical trials of cardiac amyloidosis at Northwestern Medicine. A total of 73 cardiac amyloidosis patients were identified and formed the positive instances. A random sample of 197 patients that visited cardiologists at Northwestern Medicine but were not diagnosed to have cardiac amyloidosis (based on their ICD-9 codes) formed the negative instances. The negative patients were chosen in such a way so as to maintain similarity with the positive patients but not have cardiac amyloidosis. A total of 270 patients formed our training dataset as summarized in Table 1.

Table 1. Number of instances in each category

| Positive cardiac amyloidosis patients | 73 |
|---|---|
| Negative cardiac amyloidosis patients | 197 |
| **Total** | **270** |

Next, we represented each patient as a vector using information from their EHRs stored in Northwestern Medicine Enterprise Data warehouse (NMEDW). Each patient record was then processed so that all relevant sections of the patient records which are potentially useful for creating features space were extracted. The relevant sections or feature types we have considered in this study are Demographics (such as Gender, Race, Ethnicity and Age), Diagnosis, Medication, Problems, Surgical data and Progress Notes. Demographics, Diagnosis and Medication, Problems and Surgical data are available in structured form, whereas Progress notes are only available as unstructured text. The value of each feature type is then transformed in order to be useful for machine learning classifier. Each value of Demographics, Diagnosis and Medication is treated as a binary feature. For example, if a patient has diagnosis "Anemia", we assigned a vector for "Anemia" and assigned a value of 1 for that patient and 0 for other patients. Similarly the presence of each medication such "Paracetamol" is treated as a separate feature. Certain variable such as 'age' in Demographics are numeric with a wide range of values. Such features are discretized[14] by transforming the numbers to assignment in bins or ranges. That is, rather than representing the age of the patient by its numeric value (for e.g. 43), we represent it by a range (e.g. 40-49). Processing the records this way generated, 15 Demographics features, 5002 Diagnosis features, 3521 Medication features, 2124 Problem features and 101 Surgical data features. For transforming Progress notes to informative features, we follow the widely used Bag-of-Words model. The progress notes from all patient records are first preprocessed (lowercase, removal of punctuation and stop words, normalizing numeric values to placeholders) and then tokenized to unigrams and bigrams bag of words. These are then further pruned by ignoring the very high frequency unigrams or bigrams. After the transformation, in total, we get 1766 unigram features and 1044 bigram features for this dataset. The features are summarized in Table 2.

**Table 2.** Feature types and number of features under each type

| | |
|---|---|
| Demographics Features | 15 |
| Diagnosis Features | 5002 |
| Medication Features | 3521 |
| Problem Features | 2124 |
| Surgical Features | 101 |
| Unigram Features | 1766 |
| Bigram Features | 1044 |
| **Total** | **13573** |

On this dataset of 223 instances and 8011 features several machine learning classifiers which we describe next are trained.

The machine learning classification algorithms we have considered in our study are K Nearest Neighbors,[15,16] Support Vector Machines,[17,18] Decision Trees,[19] Random Forests,[20] AdaBoost,[21] and Naïve Bayes[22,23]. K-nearest neighbors (KNN) is an instance based learning algorithm that stores all the training instances and classifies the new cases based on the distance functions. K-NN works on the principle of classifying the new test case by a majority vote, with the test case being assigned to the class most common amongst its K nearest neighbors. Each point in the neighborhood may be assigned equal importance or given some sort of weight such as distance (larger the distance smaller the weight/importance).

Support Vector Machine (SVM) performs classification by finding the hyperplane that maximizes the margin between the two classes. The instances that define the hyperplane are called support vectors. SVM is high performing machine learning classifier due to the fact that if the data is linearly separable, SVM would produce a hyperplane that completely separates the vectors into two classes. However, in real life scenario perfect separation may not be possible. In such a case we can use different kernel function to project the data point into a different feature space where data may be linearly separable. Kernel function can be viewed as feature extraction techniques which take the existing features as input, performs the transformation and output a new feature space.

Naïve Bayes (NB) algorithm is based on the Bayes theorem with assumption that the features are independent of each other. NB model is easy to develop with no parameter estimation. Despite its simplicity, NB model is widely used and outperforms other sophisticated machine learning algorithms such as SVM. NB provides a method of calculating posterior probability of class given the prior probability and learning the likelihood of features given the class.

Decision trees (DT) builds classification models using a tree like structure. It works on the principle of divide and conquer breaking the dataset at each feature according to the best split and at the same time incrementally building the tree. The final result of the algorithm is a tree with decision node and leaf nodes. Decision node makes a certain conclusion depending upon the feature and has two or more branches. Leaf node does the final classification.

Random Forest (RF) and AdaBoost are two ensemble algorithms which we have also used in our experiments. The principle behind working of RF algorithm is to construct many decision trees during the training and classifies the test case based on the mode of the classes given by individual trees. AdaBoost is another ensemble methods in which various other machine learning classifiers or also weak learners are combined to give a weighted sum that represents the output of the final classifier. It begins by fitting a classifier first on the data and then learning more classifiers by adjusting the weights of the previously misclassified data points.

In addition, there are also a variety of hyper-parameters that were tuned in tandem for each of these classification algorithms. Table 3 lists the various algorithms and the corresponding hyper-parameters and their value ranges involved with each algorithm. We perform grid search[24] technique which is a brute-force exhaustive method to search for the best parameters through the manually specified subset of the hyper-parameter space of a learning algorithm.

Table 3. Classification Algorithms used, their hyper-parameters and range

| Classification Algorithm | Hyper-parameter | Range of Value searched over |
|---|---|---|
| K-Nearest Neighbor | K – Number of nearest neighbors<br>Weights – weight of points in neighborhood | K – {1, 2, 3}<br>Weights – {Uniform, Distance} |
| Support Vector Machines | C – Loss regularization tradeoff<br>Kernel – Kernel type<br>Gamma – kernel coefficient | C – {0.01, 0.1, 1, 10, 100}<br>Kernel – {linear, rbf}<br>Gamma – {1e-3, 1e-4} |
| Naïve Bayes | None | |
| Decision Trees | Criterion – Quality measure of the split | Criterion – {gini, entropy} |
| Random Forests | N-Estimator – Number of trees in forest<br>Criterion – Quality measure of the split | N-Estimator – {10, 15, 100}<br>Criterion – {gini, entropy} |
| AdaBoost | N-Estimator – Maximum number of estimators | N-Estimator – {10, 15, 100} |

In addition to various algorithms and models, we also employed feature selection methods to improve upon the accuracy of the classification models. First, features with very low variance are removed. That is, the features whose 85% of values are either '0' or '1' are removed since they do not give much information about either category. We then used L1 regularization[25] to do more feature selection and thereby identify top features to identify those that play decisive roles in classifying the patient into each category. L1 regularization adds a weight penalty to the loss function. Since each non-zero weight adds to the penalty, it forces weak features to have zero weight. Thus, this produces a sparse solution by intrinsically performing feature selection.

To evaluate the system, we perform tenfold cross-validation and report the average F1 score[26] (harmonic mean of precision and recall).

**Results**

Table 4. Optimal Parameters for each machine learning and average F1 score

| Algorithm | Configuration | Average 10 cross validation F1 score |
|---|---|---|
| K Nearest Neighbor | K – 1, criterion – 'uniform' | 0.42 |
| Naïve Bayes | | 0.71 |
| SVM | Kernel – rbf, C -100, gamma - 0.001 | 0.93 |
| Decision Tree | Criterion – gini | 0.94 |
| Random Forest | N-estimator - 15, Criterion – gini | 0.94 |
| **AdaBoost** | **N-estimator - 15** | **0.97** |

Table 4 shows the optimal configuration for each classification algorithm used as measured by average tenfold cross-validation F1 score. Since the number of positive instances in the dataset is very small, the number of nearest neighbors (KNN) take the least possible value 1 (K in KNN). KNN and NB gives a very low F1 score as compared to other complex algorithms since they are sensitive to outliers and also do not perform adequate feature selection. These algorithms give equal importance to all the features, which leads to poor performance. On the other hand, SVM, DT and Ensemble methods give a large F1-score due to complexity in models and ability to intrinsically performing feature selection. Also, as expected, ensemble method AdaBoost outperforms rest of the algorithms since it combines the output of other weak learners to arrive at the final hypothesis.

**Table 5.** Top Features or Factors from feature selection

| |
|---|
| Age range of 80-89 |
| Age range of 70-79 |
| Diagnosis of cardiac arrest |
| Diagnosis of chest pain |
| Diagnosis of congestive heart failure |
| Diagnosis of hypertension |
| Diagnosis of prim open angle glaucoma |
| Diagnosis of shoulder arthritis |
| Medication of Doxercalciferol |
| Medication of Loratadine 10mg |
| Medication of Levothyroxine Sodium 25mcg |
| Medication of 008278-ZZceFAZolin Sodium |
| Medication of Sodium Chloride 0.9% |

We list some of the important features obtained after performing feature selection in Table 5. It can be seen that patients in age range 70-90 are more likely to have cardiac amyloidosis than patients of other ages. Also, presence of other cardiac diagnosis such as congestive heart failure and chest pain as important features shows that cardiac amyloidosis are comorbid with other heart problems. Some of the medications are also suggestive of this hypothesis. However, this requires further analysis and inputs from multiple cardiologists.

The accuracy measures on re-running the machine learning algorithms using only the features filtered from feature selection technique are listed in Table 6. All the algorithms give nearly high and similar average cross-validation F1 score. As anticipated, there is also rise in scores of KNN and NB due to removal of noisy insignificant features.

**Table 6.** Accuracy Measures after feature selection

| Algorithm | Average 10 cross validation F1 score |
|---|---|
| K Nearest Neighbor | 0.95 |
| Naïve Bayes | 0.974 |
| SVM | 0.976 |
| Decision Tree | 0.974 |
| **Random Forest** | **0.981** |
| AdaBoost | 0.975 |

**Discussion**

This work is very relevant to an emerging direction in clinical informatics research focusing on developing patient similarity measures derived from EHR data for application in a variety of areas.[27-36] For example, Zhang et al used patient similarity and drug similarity for personalized medicine.[35] Wang et al attempted to implement treatment recommendations and Sanders et al used similarity measure for visualizing which patients are similar.[27,28] Li et al used topological similarity measures to identify subgroups of type 2 diabetes patients.[31] Huang et al applied such metrics for analyzing clinical pathways and Lee et al and Ng et al applied for predicting mortality for individual patients.[32,36] There are systems such as PSF that are evaluating novel metric learning approaches and those that are using external knowledge resources such as PubMed to improved patient similarity measures.[33,34]

We also used Apache cTAKES[37] to extract more features from unstructured Progress notes. However, since we are already accomplishing a very high average cross-validation F1 score and the unigram and bigram features that are also obtained from progress notes are not very predictive, we decided not to currently use them in further experiments. These would have only further escalated the curse of dimensionality[38] and will not have contributed significantly to increase the accuracy of the models. However, we might implement NLP-based features as we are expanding our algorithm to other rare diseases.

In spite of the high accuracy of our system, there is a limitation. In our training set, the instances are moderately balanced, i.e., positive and negative instances are 1:3 in ratio. However, as we have mentioned in the Introduction, cardiac amyloidosis is a rare disease. The ratio in real scenario will be closer to 1:1000. If we directly use the models produced through our methods, we may get a large number of false positives and thereby low precision rate because of failure to generalize and the sample data not being true representative of real world data[39]. To counter this, we need to do preprocessing using high recall rules to filter out the very obvious negative instances. The goal of these rules will be to reduce the skewness of the real world data. After the filtering step, our models can be applied on the resulting dataset. These high recall rules can be implemented using the factors identified through our methods as most predictive in Table 5. We can also employ other feature selection techniques such as wrapper methods, tree based induction methods[40] or use association mining[41,42] for doing. We consider this beyond the scope of this initial work.

**Conclusion**

We have developed a machine learning system to automatically identify cardiac amyloidosis patients using EHR data. We performed data processing, parameter estimation, feature selection and model selection to achieve a very high tenfold cross-validation F1 score of 0.98 using Random Forest, an ensemble machine learning classifier. We have also reported some of the most predictive features. However, our system needs to be validated for its operational value across an entire healthcare system by executing it on system-wide EHR data for identifying rare disease patients and manually verifying the accuracy.

**Acknowledgement**
This project was partially supported by National Library of Medicine (grant R00LM011389).

**References**

1. Sox HC, Higgins MC. Medical decision making: ACP Press; 1988.
2. Sullivan J. Subject Recruitment and Retention: Barrier to Success. Applied Clinical Trials 2004.
3. Zhang M, Zhu C, Jacomy A, Lu LJ, Jegga AG. The orphan disease networks. The American Journal of Human Genetics 2011;88:755-66.
4. Schieppati A, Henter JI, Daina E, Aperia A. Why rare diseases are an important medical and social issue. Lancet 2008;371:2039-41.
5. Stolk P, Willemen MJ, Leufkens HG. Rare essentials: drugs for rare diseases as essential medicines. Bulletin of the World Health Organization 2006;84:745-51.
6. Quarta CC, Kruger JL, Falk RH. Cardiac Amyloidosis. Circulation 2012;126:e178-e82.
7. Jensen PB, Jensen LJ, Brunak S. Mining electronic health records: towards better research applications and clinical care. Nat Rev Genet 2012;13:395-405.

8. Friedman CP, Wong AK, Blumenthal D. Achieving a Nationwide Learning Health System. Science Translational Medicine 2010;2.
9. Friedman C, Rigby M. Conceptualising and creating a global learning health system. Int J Med Inform 2013;82:e63-71.
10. Ma X-J, Wang Z, Ryan PD, et al. A two-gene expression ratio predicts clinical outcome in breast cancer patients treated with tamoxifen. Cancer Cell 2004;5:607-16.
11. Strom BL, Schinnar R, Jones J, et al. Detecting pregnancy use of non-hormonal category X medications in electronic medical records. Journal of the American Medical Informatics Association 2011;18:i81-i6.
12. Mathias JS, Gossett D, Baker DW. Use of electronic health record data to evaluate overuse of cervical cancer screening. Journal of the American Medical Informatics Association 2012;19:e96-e101.
13. De Pauw R, Kregel J, De Blaiser C, et al. Identifying prognostic factors predicting outcome in patients with chronic neck pain after multimodal treatment: A retrospective study. Man Ther 2015;20:592-7.
14. Liu H, Hussain F, Tan CL, Dash M. Discretization: An enabling technique. Data mining and knowledge discovery 2002;6:393-423.
15. Altman NS. An introduction to kernel and nearest-neighbor nonparametric regression. The American Statistician 1992;46:175-85.
16. Ruiz EV. An algorithm for finding nearest neighbours in (approximately) constant average time. Pattern Recognition Letters 1986;4:145-57.
17. Suykens JA, Vandewalle J. Least squares support vector machine classifiers. Neural processing letters 1999;9:293-300.
18. Cortes C, Vapnik V. Support vector machine. Machine learning 1995;20:273-97.
19. Quinlan JR. Induction of decision trees. Machine learning 1986;1:81-106.
20. Liaw A, Wiener M. Classification and regression by randomForest. R news 2002;2:18-22.
21. Rätsch G, Onoda T, Müller K-R. Soft margins for AdaBoost. Machine learning 2001;42:287-320.
22. Rish I. An empirical study of the naive Bayes classifier.  IJCAI 2001 workshop on empirical methods in artificial intelligence; 2001: IBM New York. p. 41-6.
23. McCallum A, Nigam K. A comparison of event models for naive bayes text classification.  AAAI-98 workshop on learning for text categorization; 1998: Citeseer. p. 41-8.
24. Bergstra J, Bengio Y. Random search for hyper-parameter optimization. The Journal of Machine Learning Research 2012;13:281-305.
25. Ng AY. Feature selection, L 1 vs. L 2 regularization, and rotational invariance.  Proceedings of the twenty-first international conference on Machine learning; 2004: ACM. p. 78.
26. Powers DM. Evaluation: from precision, recall and F-measure to ROC, informedness, markedness and correlation. 2011.
27. Wang Y, Tian Y, Tian LL, Qian YM, Li JS. An electronic medical record system with treatment recommendations based on patient similarity. Journal of medical systems 2015;39:55.
28. Sanders M, Winters P, Fiscella K. Preliminary validation of a scale to measure patient perceived similarity to their navigator. BMC Res Notes 2015;8:388.
29. Panahiazar M, Taslimitehrani V, Pereira NL, Pathak J. Using EHRs for Heart Failure Therapy Recommendation Using Multidimensional Patient Similarity Analytics. Stud Health Technol Inform 2015;210:369-73.
30. Ng K, Sun J, Hu J, Wang F. Personalized Predictive Modeling and Risk Factor Identification using Patient Similarity. AMIA Joint Summits on Translational Science proceedings AMIA Summit on Translational Science 2015;2015:132-6.
31. Li L, Cheng WY, Glicksberg BS, et al. Identification of type 2 diabetes subgroups through topological analysis of patient similarity. Sci Transl Med 2015;7:311ra174.
32. Lee J, Maslove DM, Dubin JA. Personalized mortality prediction driven by electronic medical data and a patient similarity metric. PLoS One 2015;10:e0127428.
33. Fei W, Sun J. PSF: A Unified Patient Similarity Evaluation Framework Through Metric Learning With Weak Supervision. IEEE journal of biomedical and health informatics 2015;19:1053-60.
34. Chan LW, Liu Y, Chan T, et al. PubMed-supported clinical term weighting approach for improving inter-patient similarity measure in diagnosis prediction. BMC Med Inform Decis Mak 2015;15:43.
35. Zhang P, Wang F, Hu J, Sorrentino R. Towards personalized medicine: leveraging patient similarity and drug similarity analytics. AMIA Joint Summits on Translational Science proceedings AMIA Summit on Translational Science 2014;2014:132-6.

36. Huang Z, Dong W, Duan H, Li H. Similarity measure between patient traces for clinical pathway analysis: problem, method, and applications. IEEE journal of biomedical and health informatics 2014;18:4-14.
37. Savova GK, Masanz JJ, Ogren PV, et al. Mayo clinical Text Analysis and Knowledge Extraction System (cTAKES): architecture, component evaluation and applications. Journal of the American Medical Informatics Association 2010;17:507-13.
38. Friedman JH. On bias, variance, 0/1—loss, and the curse-of-dimensionality. Data mining and knowledge discovery 1997;1:55-77.
39. Kotsiantis S, Kanellopoulos D, Pintelas P. Handling imbalanced datasets: A review. GESTS International Transactions on Computer Science and Engineering 2006;30:25-36.
40. Kohavi R, John GH. Wrappers for feature subset selection. Artificial intelligence 1997;97:273-324.
41. Ceglar A, Roddick JF. Association mining. ACM Computing Surveys (CSUR) 2006;38:5.
42. Agrawal R, Srikant R. Fast algorithms for mining association rules.  Proc 20th int conf very large data bases, VLDB; 1994. p. 487-99.